\documentclass{article} % For LaTeX2e
\pdfoutput=1
\usepackage{iclr2018_conference,times}
\usepackage{hyperref}
\usepackage{url}
\usepackage{graphicx}
\usepackage{amssymb,amsmath,bm}
\usepackage{textcomp}
\usepackage{tikz}
\usepackage{caption}
\usepackage{subcaption}
\usepackage{booktabs}
\usepackage{lineno}
\usepackage{array}
\usepackage{etoolbox,siunitx}
\robustify\bfseries

\usetikzlibrary{arrows}
\usetikzlibrary{calc}
\DeclareSymbolFont{extraup}{U}{zavm}{m}{n}
\DeclareMathSymbol{\varheart}{\mathalpha}{extraup}{86}
\DeclareMathSymbol{\vardiamond}{\mathalpha}{extraup}{87}

\title{Twin Networks: Matching the Future \\
for Sequence Generation}

\author{Dmitriy Serdyuk,$^\textbf{*}$\,${}^\vardiamond$ Nan Rosemary Ke,${}^{\textbf{*}}\,^{\vardiamond\,\ddagger}$ Alessandro Sordoni$^\varheart$
\vspace{0.1cm}\\
\textbf{Adam Trischler,}$^\varheart$ \textbf{Chris Pal}$^\clubsuit{}^\vardiamond$ \textbf{\&} \textbf{Yoshua Bengio}$^{\P\,\vardiamond}$ \vspace{5mm} \\
$^\vardiamond$ Montreal Institute for Learning Algorithms (MILA), Canada \\
$^\varheart$ Microsoft Research, Canada \\
$^\clubsuit$ Ecole Polytechnique, Canada \\
$^\P$ CIFAR Senior Fellow \\
$^\ddagger$ Work done at Microsoft Research \\
$^\textbf{*}$ \textbf{Authors contributed equally} \\
\texttt{serdyuk@iro.umontreal.ca}, \texttt{rosemary.nan.ke@gmail.com}
\vspace{1.5cm}
}

\iclrfinalcopy % Uncomment for camera-ready version, but NOT for submission.

\begin{document}

\maketitle
% because of phantom author
\vspace*{-1.6cm}
\begin{abstract}
We propose a simple technique for encouraging generative RNNs to plan ahead.
We train a ``backward'' recurrent network to generate a given sequence in reverse order, and we encourage states of the forward model to predict cotemporal states of the backward model.
The backward network is used only during training, and plays no role during sampling or inference.
We hypothesize that our approach eases modeling of long-term dependencies by implicitly forcing the forward states to hold information about the longer-term future (as contained in the backward states).
We show empirically that our approach achieves 9\% relative improvement for a speech recognition task, and achieves significant improvement on a COCO caption generation task.
\end{abstract}

\section{Introduction}
\label{sec:intro}
Recurrent Neural Networks (RNNs) are the basis of state-of-art models for generating sequential data such as text and speech. RNNs are trained to generate sequences by predicting one output at a time given all previous ones, and excel at the task through their capacity to remember past information well beyond classical $n$-gram models~\citep{bengio1994learning,hochreiter1997long}.
More recently, RNNs have also found success when applied to conditional generation tasks such as speech-to-text~\citep{NIPS2015_5847,chan2015listen}, image captioning~\citep{xu2015show} and machine translation~\citep{sutskever2014sequence,bahdanau2014neural}.

RNNs are usually trained by \emph{teacher forcing}: at each point in a given sequence, the RNN is optimized to predict the next token given all preceding tokens.
% , rather than all previously \emph{generated} tokens (from the model)
This corresponds to optimizing one-step-ahead prediction. As there is no explicit bias toward planning in the training objective, the model may prefer to focus on the most recent tokens instead of capturing subtle long-term dependencies that could contribute to global coherence. Local correlations are usually stronger than long-term dependencies and thus end up dominating the learning signal. The consequence is that samples from RNNs tend to exhibit local coherence but lack meaningful global structure. This difficulty in capturing long-term dependencies has been noted and discussed in several seminal works~\citep{hochreiter1991untersuchungen,bengio1994learning,hochreiter1997long,pascanu2013difficulty}.

Recent efforts to address this problem have involved augmenting RNNs with external memory~\citep{dieng2016topicrnn,grave2016improving,gulcehre2017memory}, with unitary or hierarchical architectures~\citep{arjovsky2016unitary,serban2017hierarchical}, or with explicit planning mechanisms~\citep{tris2017}. Parallel efforts aim to prevent overfitting on strong local correlations by regularizing the states of the network, by applying dropout or penalizing various statistics~\citep{moon2015rnndrop,zaremba2014recurrent,gal2016theoretically,krueger2016zoneout,merity2017regularizing}.

%Recent efforts involved augmenting RNNs with external memory~\citep{mikolov2012context,dieng2016topicrnn,grave2016improving}, with unitary or hierarchical architectures~\citep{koutnik2014clockwork,arjovsky2016unitary,serban2017hierarchical} or with explicit planning mechanisms~\citep{li2017learning,tris2017}. In parallel, other efforts aim to avoid over-fitting strong local correlations during training by regularizing the states of the network,~i.e. by applying dropout on the hidden states or inputs~\citep{moon2015rnndrop,zaremba2014recurrent,gal2016theoretically,krueger2016zoneout} or by penalizing the norm of the hidden states~\citep{merity2017regularizing}.

\begin{figure}
      \centering
  \begin{tikzpicture}[->,thick]
\scriptsize
\tikzstyle{main}=[circle, minimum size = 7mm, thin, draw =black!80, node distance = 12mm]

\foreach \name in {1,...,4}
    \node[main, fill = white!100] (y\name) at (\name*1.5,3.5) {$x_\name$};

\foreach \name in {1,...,4}
    \node[main, fill = white!100] (hf\name) at (\name*1.5,1.5) {$h^f_\name$};
\foreach \name in {1,...,4}
    \node[main, fill = white!100,draw=orange] (hb\name) at (\name*1.5,0) {$h^b_\name$};
\foreach \h in {1,...,4}
       {
        \draw[<->,draw=orange,dashed] (hf\h) to [bend right=45] node[midway,left] {$L_\h$} (hb\h)  {};
        
        \path (hf\h) edge [bend left] (y\h);
       }
\foreach \current/\next in {1/2,2/3,3/4} 
       {
        \path (hf\current) edge (hf\next);
        \path[draw=orange] (hb\next) edge (hb\current);
       }
\foreach \h in {1,...,4}
       {
        \path (hb\h) edge [draw=orange,bend right] (y\h);     
       }
    %\node[main] (G-\name) at (\x,0) {$\name$};
\end{tikzpicture}
      \caption{The forward and
        the backward networks predict the sequence $s = \{x_1, ..., x_4\}$ independently.
        The penalty matches the forward (or a parametric function of the forward) and the backward hidden states. The forward network receives the gradient signal from the log-likelihood objective as well as $L_t$  between states that predict the same token. The backward network is trained only by maximizing the data log-likelihood. During the evaluation part of the network colored with orange is discarded. The cost $L_t$ is either a Euclidean distance or a learned metric $||g(h_t^f) - h_t^b||{}_2$ with an affine transformation $g$. Best viewed in color.}
      \label{fig:twin}
  \end{figure}
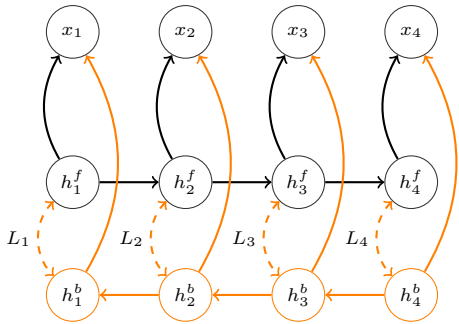

In this paper, we propose \emph{TwinNet},\footnote{
%The preliminary version of this paper was presented at a workshop. 
The source code is available at \url{https://github.com/dmitriy-serdyuk/twin-net/}.} a simple method for regularizing a recurrent neural network that encourages modeling those aspects of the past that are predictive of the long-term future.
%In addition to predicting the next token in the sequence,  the hidden state to contain information about the whole future of the sequence.
Succinctly, this is achieved as follows: in parallel to the standard forward RNN, we run a ``twin'' backward RNN (with no parameter sharing) that predicts the sequence in reverse, and we encourage the hidden state of the forward network to be close to that of the backward network used to predict the same token.
Intuitively, this forces the forward network to focus on the past information that is useful to predicting a specific token and that is \emph{also} present in and useful to the backward network, coming from the future (Fig.~\ref{fig:twin}).

In practice, our model introduces a regularization term to the training loss.
This is distinct from other regularization methods that act on the hidden states either by injecting noise~\citep{krueger2016zoneout} or by penalizing their norm~\citep{krueger2015regularizing,merity2017regularizing}, because we formulate explicit auxiliary targets for the forward hidden states: namely, the backward hidden states.
The activation regularizer (AR) proposed by~\cite{merity2017regularizing}, which penalizes the norm of the hidden states, is equivalent to the TwinNet approach with the backward states set to zero.
Overall, our model is driven by the intuition (a) that the backward hidden states contain a summary of the future of the sequence, and (b) that in order to predict the future more accurately, the model will have to form a better representation of the past.
We demonstrate the effectiveness of the TwinNet approach experimentally, through several conditional and unconditional generation tasks that include speech recognition, image captioning, language modelling, and sequential image generation. 
% Our model is specifically focused on conditional generation to model the conditional probability of a sequence $s$ given context information $c$, $p(x|c)$, which contains much less entropy than unconditional generation. To be more specific, this is to avoid the impossible challenge of making the first few tokens containing all the information of the future, which could have very high entropy in unconditional generative models. In this paper, we evaluate our model in the setting of conditional generation for speech recognition, image captioning, and unconditional generation for language modelling.
To summarize, the contributions of this work are as follows:
\begin{itemize}
    \item We introduce a simple method for training generative recurrent networks that regularizes the hidden states of the network to anticipate future states (see Section~\ref{sec:model});
    \item The paper provides extensive evaluation of the proposed model on multiple tasks and concludes that it helps
    training and regularization for conditioned generation (speech recognition, image captioning) and for the unconditioned case (sequential MNIST, language modelling, see Section~\ref{sec:experiments});
    \item For deeper analysis we visualize the introduced cost and observe that it  negatively correlates with the word frequency (more surprising words have higher cost).
\end{itemize}

\pagebreak
\section{Model}
\label{sec:model}
Given a dataset of sequences $\mathcal{S} = \{s^1, \ldots, s^n\}$, where each $s^k = \{x_1, \ldots, x_{T_k}\}$ is an observed sequence
of inputs $x_i \in \mathcal{X}$, we wish to estimate a density $p(s)$ by maximizing the log-likelihood of the observed data
$\mathcal{L} = \sum_{i=1}^n \log p(s^i)$. Using the chain rule, the joint probability over a sequence $x_1, \ldots, x_T$ decomposes as: 
\begin{equation}
    p(x_1, \ldots, x_T) = p(x_1)p(x_2|x_1)... = \prod_{t=1}^{T} p(x_t | x_{1}, \ldots, x_{t-1}).
\end{equation}
This particular decomposition of the joint probability has been widely used in language modeling~\citep{bengio2003neural,mikolov2010recurrent} and speech recognition~\citep{bahl1983maximum}.
A recurrent neural network is a powerful architecture for approximating this conditional probability. At each step, the RNN updates a hidden state $h_t^f$, which iteratively summarizes the inputs seen up to time $t$:
\begin{equation}
h^f_t = \Phi_f(x_{t-1}, h_{t-1}^f),
\end{equation}
where $f$ symbolizes that the network reads the sequence in the forward direction, and $\Phi_f$ is typically a non-linear function, such as a LSTM cell~\citep{hochreiter1997long} or a GRU~\citep{cho2014learning}. Thus, $h^f_t$ forms a representation summarizing information about the sequence's past. The prediction of the next symbol $x_t$ is performed using another non-linear transformation on top of $h^f_t$,~i.e. $p_f(x_t|x_{<t}) = \Psi_f(h^f_t)$, which is typically a linear or affine transformation (followed by a softmax when $x_t$ is a symbol). The basic idea of our approach is to encourage $h^f_t$ to contain information that is useful to predict $x_t$ and which is also compatible with the upcoming (future) inputs in the sequence. To achieve this, we run a twin recurrent network that predicts the sequence in reverse and further require the hidden states of the forward and the backward networks to be close. The backward network updates its hidden state according to:
\begin{equation}
h^b_t = \Phi_b(x_{t+1}, h_{t+1}^b),
\end{equation}
and predicts $p_b(x_t | x_{>t}) = \Psi_b(h^b_t)$ using information only about the future of the sequence. Thus, $h^f_{t}$ and $h^b_{t}$ both contain useful information for predicting $x_t$, coming respectively from the past and future. Our idea consists in penalizing the distance between forward and backward hidden states leading to the same prediction. For this we use the Euclidean distance (see Fig.~\ref{fig:twin}):
\begin{equation}
L_t(s) = \| g(h_t^f) - h_t^b \|_2,
\label{eq:parametric}
\end{equation}
where the dependence on $x$ is implicit in the definition of $h_t^f$ and $h_t^b$.
The function $g$ adds further capacity to the model and comes from the class of parameterized affine transformations.
Note that this class includes the identity tranformation.
As we will show experimentally in Section~\ref{sec:experiments}, a learned affine transformation gives more flexibility to the model and leads to better results. This relaxes the strict match between forward and backward states, requiring just that the forward hidden states are predictive of the backward hidden states.\footnote{
  Matching hidden states is equivalent to matching joint distributions factorized in two different ways, since a given state
  contains a representation of all previous states for generation of all later states and outputs. For comparison, we
  made several experiments matching outputs of the forward and backward networks rather than their hidden states, which is equivalent to matching 
$p(x_t | x_{<t})$ and $p(x_t | x_{>t})$ separately for every $t$. None of these experiments converged.}

%RNNs factorize the probability of the sequence as:
%\begin{equation*}
%\stack{p}{\rightarrow}(\mathbf{x}) = p(x_1) p(x_2 | x_1) p(x_3|x_2, x_1) \ldots %p(x_N|x_{N-1}\ldots x_1),
%\end{equation*}
%We modify the architecture by adding a second decoder with attention which produces characters but running backwards instead of forward as usual. The forward network models the factorized probability
%while the backward network models the same joint probability but factorized differently, as
%\begin{equation}
%      p(y) = p(y_N) p(y_{N-1} | y_N) \ldots p(y_1|y_2\ldots y_N),
%  \end{equation}
%  where $y$ is the output vector of length $N$. The hidden states summarize the right side of the conditioning bar.
%  For the forward network, this is the past up to this point and for the backward network this it the future.
%  Under the assumption that the hidden states are normally distributed, the L2 loss corresponds to matching
%  $h^f(y_{i-1}, y_{i-2}, \ldots)$ and $h^b(y_{i+1}, y_{i+2})\ldots$. Figure~\ref{fig:twin} demonstrates
%  the network architecture.

The total objective maximized by our model for a sequence $s$ is a weighted sum of the forward and backward log-likelihoods minus the penalty term, computed at each time-step:
\begin{equation}
\mathcal{F}(s) = \sum_{t} \log p_{f}(x_t | x_{<t}) + \log p_{b}(x_t | x_{>t}) - \alpha\, L_t(s),
\end{equation}
%where $L_{f}$ and $L_{b}$ correspond to the cross-entropy of the forward RNN and the backward RNN
%respectively with their hidden states decoded as $h_i^f$ and $h_i^b$.
where $\alpha$ is an hyper-parameter controlling the importance of the penalty term. In order to provide a more stable learning
signal to the forward network, we only propagate the gradient of the penalty term through the forward network. That is, we avoid
co-adaptation of the backward and forward networks. During sampling and evaluation, we discard the backward network.
%In general, it would be possible to adopt a different inference strategy.

The proposed method can be easily extended to the conditional generation case. The forward hidden-state transition is modified to
\begin{equation}
    h^{f}_t = \Phi_f\left(x_{t-1}, \left[h_{t-1}^{f}, c\right]\right),
\end{equation}
where $c$ denotes the task-dependent conditioning information, and similarly for the backward RNN.

\section{Related Work}
\label{sec:related}

Bidirectional neural networks~\citep{schuster1997bidirectional} have been used as powerful feature extractors for sequence tasks. The hidden state at each time step includes both information from the past and the future. For this reason, they usually act as better feature extractors than the unidirectional counterpart and have been successfully used in a myriad of tasks,~e.g. in machine translation~\citep{bahdanau2015}, question answering~\citep{chen2017reading} and sequence labeling~\citep{ma2016end}. However, it is not straightforward to apply these models to sequence generation~\citep{zhang2018asynchronous} due to the fact that the ancestral sampling process is not allowed to look into the future. In this paper, the backward model is used to regularize the hidden states of the forward model and thus is only used during training. Both inference and sampling are strictly equivalent to the unidirectional case.

Gated architectures such as LSTMs~\citep{hochreiter1997long} and GRUs~\citep{chung2014} have been successful in easing the modeling of long term-dependencies: the gates indicate time-steps for which the network is allowed to keep new information in the memory or forget stored information.~\cite{graves2014neural,dieng2016topicrnn,grave2016improving} effectively augment the memory of the network by means of an external memory. Another solution for capturing long-term dependencies and avoiding gradient vanishing problems is equipping existing architectures with a hierarchical structure~\citep{serban2017hierarchical}. Other works tackled the vanishing gradient problem by making the recurrent dynamics unitary~\citep{arjovsky2016unitary}. In parallel, inspired by recent advances in ``learning to plan'' for reinforcement learning~\citep{silver2016predictron,tamar2016value}, recent efforts try to augment RNNs with an explicit planning mechanism~\citep{tris2017} to force the network to commit to a plan while generating, or to make hidden states predictive of the far future~\citep{li2017learning}.

Regularization methods such as noise injection are also useful to shape the learning dynamics and overcome local correlations to take over the learning process. One of the most popular methods for neural network regularization is dropout~\citep{srivastava2014dropout}.
Dropout in RNNs has been proposed in~\citep{moon2015rnndrop}, and was later extended in~\citep{semeniuta2016recurrent,gal2016theoretically}, where recurrent connections are dropped at random. Zoneout~\citep{krueger2016zoneout} modifies the hidden state to regularize the network by effectively creating an ensemble of different length recurrent networks.
% This technique randomly skips recurrent connections.
%This is equivalent to creating an ensemble of different length recurrent networks.
%Other methods proposed to help with regularization of RNN training is to constraint the hidden states in RNNs.
\cite{krueger2015regularizing} introduce a ``norm stabilization'' regularization term that ensures that the consecutive hidden states of an RNN have similar Euclidean norm. Recently,~\cite{merity2017regularizing} proposed a set of regularization methods that achieve state-of-the-art on the Penn Treebank language modeling dataset. Other RNN regularization methods include the weight noise~\citep{graves2011practical}, gradient clipping~\citep{pascanu2013difficulty} and gradient noise~\citep{neelakantan2015adding}.

% In this setting, a RNN encoder produces a conditioning vector that is input at each time-step to the decoder RNN. More advanced models take advantage of content-based soft attention~\citep{bahdanau2015,xu2015show}, which provides with a different conditioning vector at each step during the decoding process.
% The decoder model for an attention model is a generative recurrent neural network which has as additional input the convex combination of \emph{contexts} (outputs of the encoder), each corresponding to a different focus of attention. 
% The soft attention mechanism assigns different weights to each context vector providing their convex combination to the decoder.

\section{Experimental Setup and Results}
\label{sec:experiments}
We now present experiments on conditional and unconditional sequence generation, and analyze the results in an effort to understand the performance gains of TwinNet. First, we examine conditional generation tasks such as speech recognition and image captioning, where the results show clear improvements over the baseline and other regularization methods. Next, we explore unconditional language generation, where we find our model does not significantly improve on the baseline. Finally, to further determine what tasks the model is well-suited to, we analyze a sequential imputation task, where we can vary the task from unconditional to strongly conditional.

\subsection{Speech Recognition}
\begin{table}[t]
    \centering
    \caption{Average character error rate (CER, \%) on WSJ dataset decoded with the beam size 10. We compare the attention model for speech
    recognition~\citep[``Baseline,''][]{bahdanau2015end-to-end}; the regularizer proposed by~\cite{krueger2015regularizing} (``Stabilizing norm''); penalty on the L2 norm of the forward states~\citep{merity2017regularizing} (``AR''), which is equivalent to TwinNet when all the hidden states of the backward network are set to zero. We report the results of our model (``TwinNet'') both with $g = I$, the identity mapping, and with a learned $g$.}
    \label{tab:speech_results}
    \begin{tabular}{lrr}
    \toprule
    \textbf{Model} &  
       \textbf{Test CER} & \textbf{Valid CER} \\
    \midrule
    Baseline & 6.8 & 9.0 \\
    Baseline + Gaussian noise & 6.9 & 9.1\\
    Baseline + Stabilizing Norm &  6.6 & 9.0 \\
    Baseline + AR &  6.5 & 8.9 \\
    Baseline + TwinNet ($g = I$) & 6.6 & 8.7 \\
    Baseline + TwinNet (learnt $g$) & {\bf 6.2} &{\bf 8.4} \\
    \bottomrule
    \end{tabular}
\end{table}
We evaluated our approach on the conditional generation for character-level speech recognition, where the model is trained to convert the speech audio signal to the sequence of characters. The forward and backward RNNs are trained as conditional generative models with soft-attention~\citep{NIPS2015_5847}. The context information $c$ is an encoding of the audio sequence and the output sequence $s$ is the corresponding character sequence.
We evaluate our model on the Wall Street Journal (WSJ) dataset closely following the setting described in~\cite{bahdanau2015end-to-end}. We use 40 mel-filter bank features with delta and delta-deltas with their energies as the acoustic inputs to the model, these features are generated according to the Kaldi s5 recipe~\citep{povey2011kaldi}. The resulting input feature dimension is $123$. 

We observe the Character Error Rate (CER) for our validation set, and we early stop on the best CER observed so far. We report CER for both our validation and test sets. For all our models and the baseline, we follow the setup in \cite{bahdanau2015end-to-end} and pretrain the model for 1 epoch, within this period, the context window is only allowed to move forward. We then perform 10 epochs of training, where the context window looks freely along the time axis of the encoded sequence, we also perform annealing on the models with 2 different learning rates and 3 epochs for each annealing stage. We use the AdaDelta optimizer for training. We perform a small hyper-parameter search on the weight $\alpha$ of our twin loss, $\alpha \in \{2.0, 1.5, 1.0, 0.5, 0.25, 0.1\}$, and select the best one according to the CER on the validation set.\footnote{The best hyperparameter was 1.5.} 

\paragraph{Results} We summarize our findings in Table~\ref{tab:speech_results}. Our best performing model shows relative improvement of 12\% comparing to the baseline. We found that the TwinNet with a learned metric (learnt $g$) is more effective than strictly matching forward and hidden states. In order to gain insights on whether the empirical usefulness comes from using a backward recurrent network, we propose two ablation tests. For ``Gaussian Noise,'' the backward states are randomly sampled from a Gaussian distribution, therefore the forward states are trained to predict white noise. For ``AR,'' the backward states are set to zero, which is equivalent to penalizing the norm of the forward hidden states~\citep{merity2017regularizing}. Finally, we compare the model with the ``Stabilizing Norm'' regularizer~\citep{krueger2015regularizing}, that penalizes the difference of the norm of consecutive forward hidden states. Results shows that the information included in the backward states is indeed useful for obtaining a significant improvement.

% \item The backward network to a small non-zero values, this is the same as adding a small bias $b$ to norm penalty of the forward hidden states; 
%[dima: do we have this?][Rosemary:I have this results, it all a part of norm reg, i picked the best one, i can also put it in a separate result line?]
% We experimented with the values of 1e-2 to 1e-4 for each of the bias and the variance of the Gaussian noise.  
%For the qualitative analysis we plot (Figure~\ref{fig:l2-cost}, left) the L2 cost for time-steps of a given sequence. 
%We observe that the cost has spikes at positions where the entropy of the prediction is high. 
%The TwinNet performs significantly better compared to Stablizing Norm and Normalization Regularization,

\begin{figure}[t]
       \centering
        \begin{subfigure}[b]{0.32\textwidth}
        \centering
        \includegraphics[width=\linewidth ]{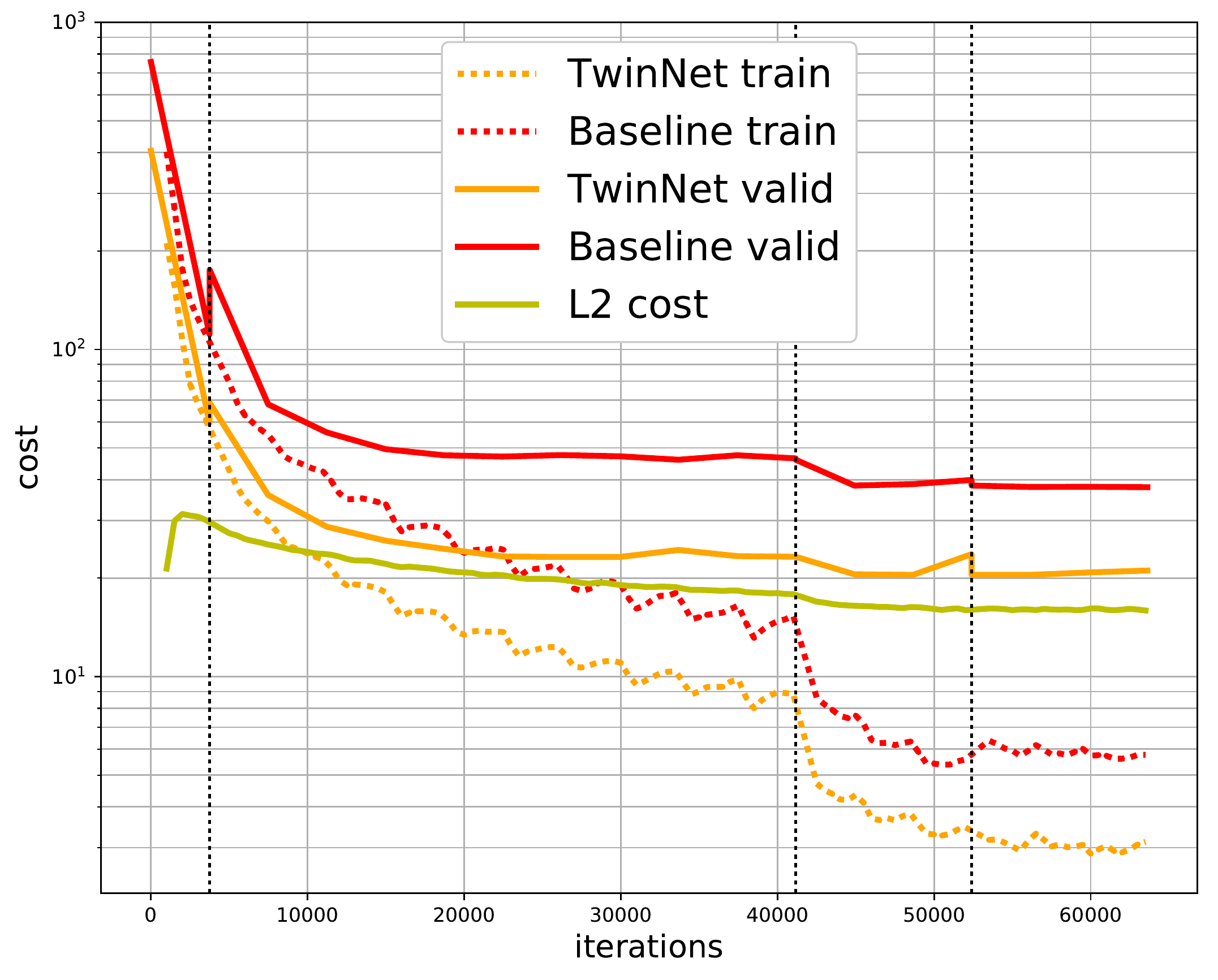}
        \caption{}
       \label{fig:train-curve}
        \end{subfigure}
        %
        %\begin{subfigure}[b]{0.32\textwidth}
        %\centering
        %\includegraphics[width=\textwidth]{figures/freq_cost.pdf}
        %\caption{{\bf (b)}:~Average L2 cost depending of frequency over the validation set. Rare words have higher cost than frequent ones.}
        %\end{subfigure}
        %
        \begin{subfigure}[b]{0.32\textwidth}
       \centering
       \includegraphics[width=\textwidth]{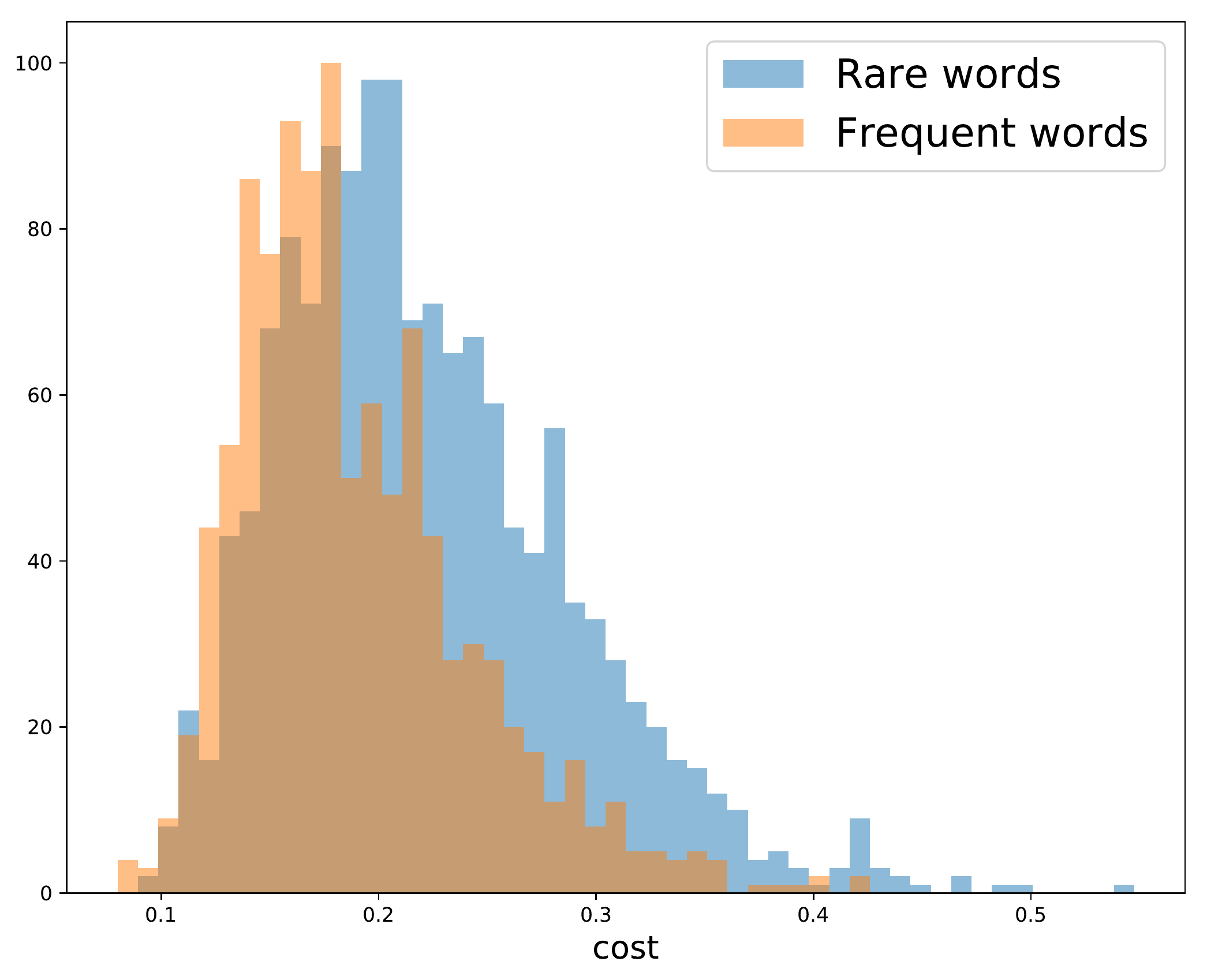}
       \caption{}
       \label{fig:hist}
        \end{subfigure}
        \begin{subfigure}[b]{0.32\textwidth}
        \centering
        \includegraphics[width=\linewidth]{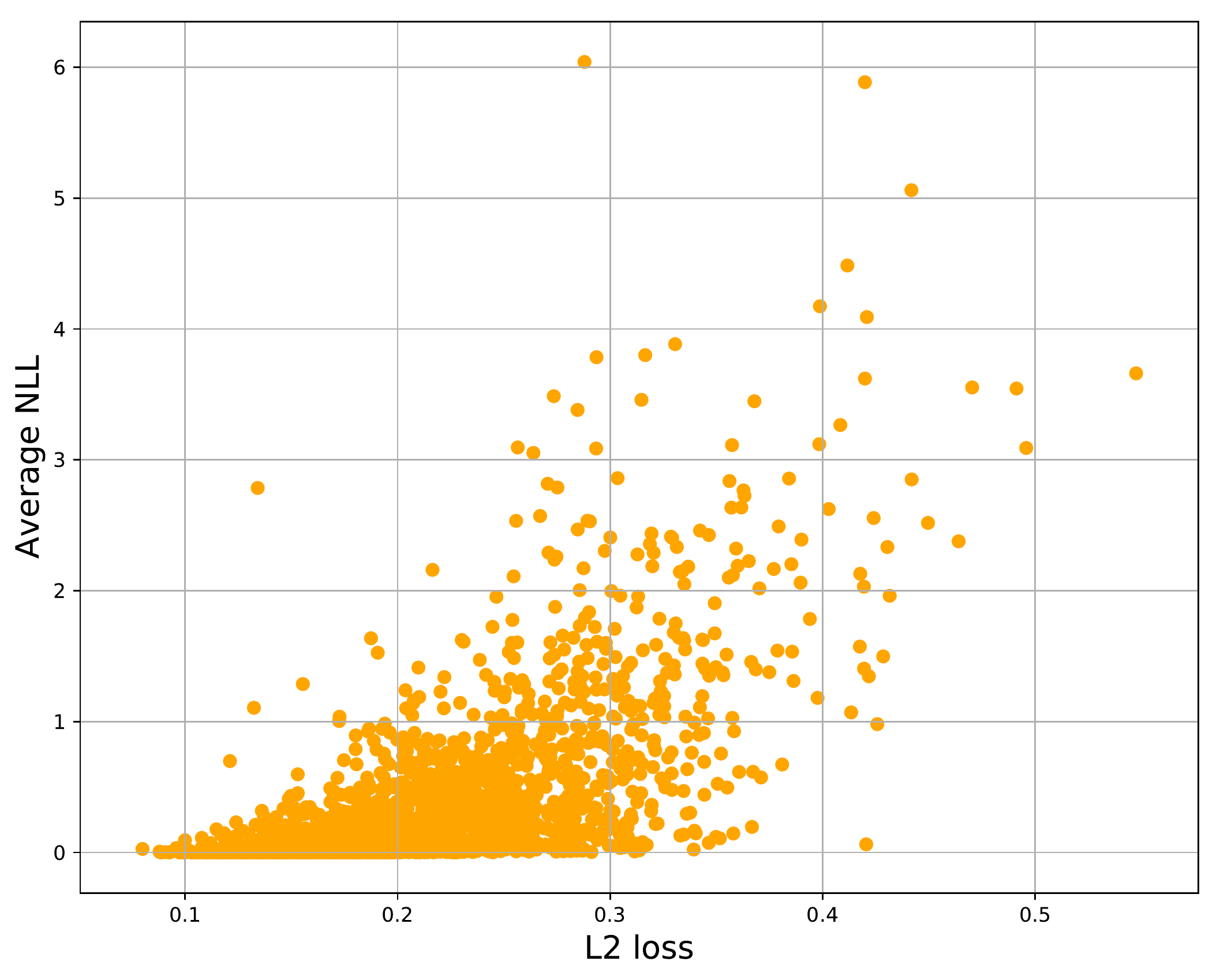}
        \caption{}
        \label{fig:l2-nll}
       \end{subfigure}
        \caption{Analysis for speech recognition experiments. {\bf (a)}:~Training curves comparison for TwinNets and the baseline network. Dotted vertical lines denote stages of pre-training, training, and two stages of annealing. The L2 cost is plotted alongside. The TwinNet converges to a better solution as well as provides better generalization. {\bf (b)}: Comparison of histograms of the cost for rare words (first 1500) versus frequent words (all other). 
         The cost is averaged over characters of a word.
         The distribution of rare words is wider and tends to produce higher L2 cost. {\bf (c)}: L2 loss vs. average cross-entropy loss.}
\end{figure}

\begin{figure}[t]
\centering
\includegraphics[width=\linewidth]{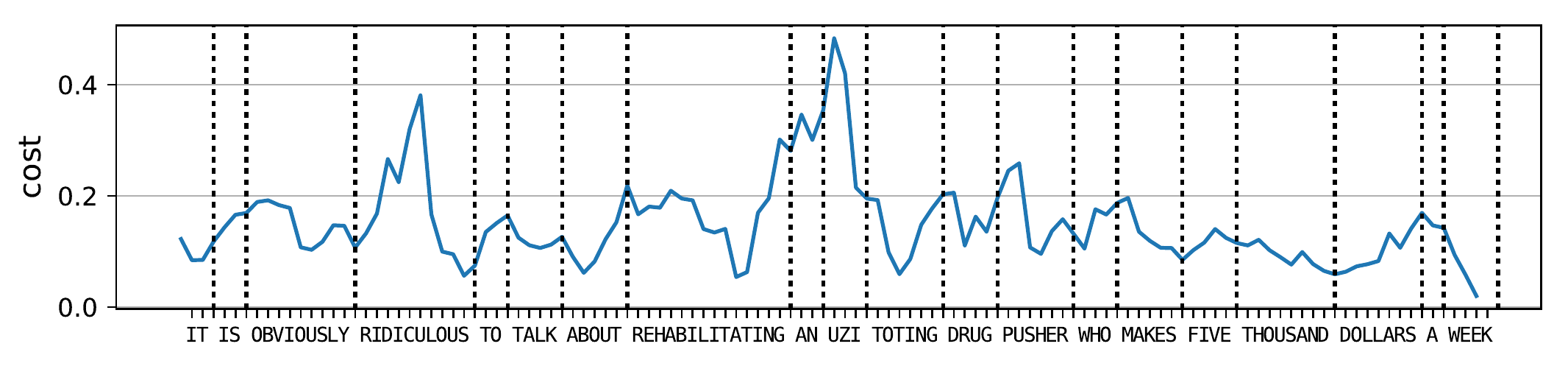}
\includegraphics[width=\linewidth]{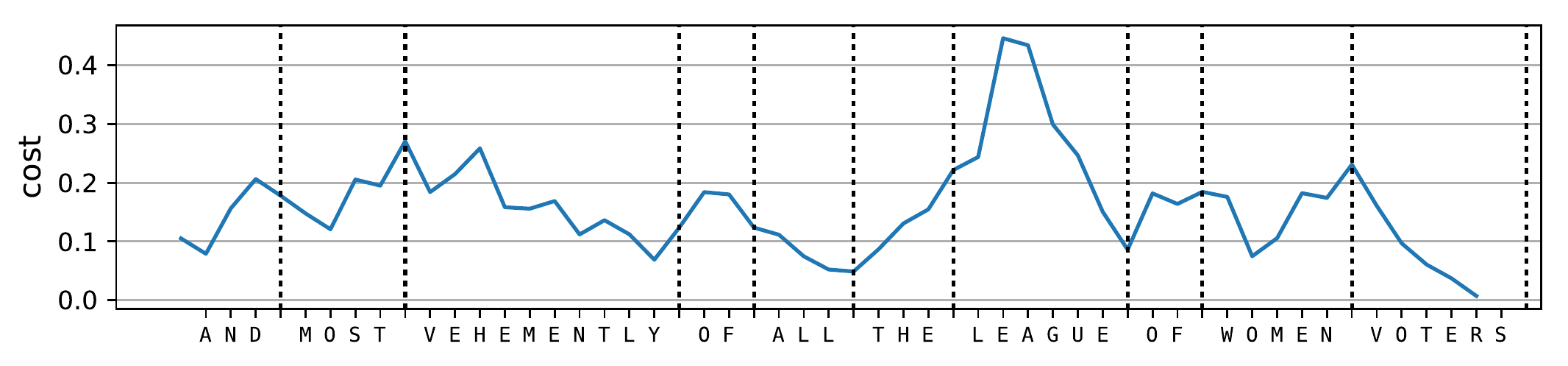}
\caption{Example of the L2 loss plotted along the time axis. Notice that spikes correspond to rare words given the acoustic information where the entropy of the prediction is high. Dotted vertical lines are plotted at word boundary positions. 
%See Appendix~\ref{appendix:asr} for more examples.
}
\label{fig:l2-cost}
\end{figure}

\paragraph{Analysis} The training/validation curve comparison for the baseline and our network is presented in Figure~\ref{fig:train-curve}.\footnote{The saw tooth pattern of both training curves corresponds to shuffling within each epoch as was previously noted  by~\cite{bottou2009curiously}.} The TwinNet converges faster than the baseline and generalizes better. The L2 cost raises in the beginning as the forward and backward network start to learn independently. Later, due to the pressure of this cost, networks produce more aligned hidden representations. Figure~\ref{fig:l2-cost} provides examples of utterances with L2 plotted along the time axis. We observe that the high entropy words produce spikes in the loss for such words as ``uzi.'' This is the case for rare words which are hard to predict from the acoustic information.
To elaborate on this, we plot the L2 cost averaged over a word depending on the word frequency. The average distance decreases with the increasing frequency. The histogram comparison  (Figure~\ref{fig:hist}) for the cost of rare and frequent words reveal that the not only the average cost is lower for frequent words, but the variance is higher for rare words. Additionally, we plot the dependency of the L2 cost cross-entropy cost of the forward network (Figure~\ref{fig:l2-nll}) to show that the conditioning also plays the role in the entropy of the output, the losses are not absolutely correlated. 
  

\subsection{Image Captioning}

We evaluate our model on the conditional generation task of  image captioning task on  Microsoft COCO dataset~\citep{lin2014microsoft}.
The MS COCO dataset covers 82,783 training images and 40,504 images for validation. Due to the lack of standardized split of training,
validation and test data, we follow Karpathy's split~\citep{karpathy2015deep, xu2015show, wang2016image}. These are 80,000 training images
and 5,000 images for validation and test.  We do early stopping based on the validation CIDEr scores and we report BLEU-1 to BLEU-4, CIDEr, and Meteor scores. To evaluate the consistency of our method, we tested TwinNet on both encoder-decoder~\citep[`Show\&Tell', ][]{vinyals2015show} and soft attention~\citep[`Show, Attend and Tell', ][]{xu2015show} image captioning models.\footnote{Following the setup in  \url{https://github.com/ruotianluo/neuraltalk2.pytorch}.}

%We used the ``Show, Attend and Tell''~\citep[SAT,][]{xu2015show} model for image captioning.\footnote{Following the setup in  \url{https://github.com/ruotianluo/neuraltalk2.pytorch}.}
We use a Resnet~\citep{he2016deep} with 101 and 152 layers pre-trained on ImageNet for image classification. The last layer of the Resned is used to extract
2048 dimensional input features for the attention model~\citep{xu2015show}. We use an LSTM with 512 hidden units for both ``Show \& Tell'' and soft attention. Both models are trained with the Adam \citep{kingma2014adam} optimizer with a learning rate of $10^{-4}$. TwinNet showed consistent improvements over ``Show \& Tell'' (Table~\ref{tab:captioning}). For the soft attention model we observe small but consistent improvements for majority of scores.

%%% report how much improvement e.g. for CIDER

%To evaluate the consistency of our method, we tested Twin Networks on both ``Show\&Tell''~\citep{vinyals2015show} and ``Show, Attend and Tell''~\citep{} image captioning models. We follow the set up in~\citep{github link}. We use  pretrained  Resnet~\citep{he2016deep} with 152 layers trained on Imagenet image classificatin  to extract features.  We then uses these extracted features for the captioning model.

%We experiment on  "Show Attend Tell" (SAT) using the COCO dataset \citep{}. The model is an encoder-decoder model with attention mechanism \citep{attention paper}. We do early stopping on the validation set based on Cider score. 

%``Show\&Tell'' is an encoder-decoder image captioning model, and "Show Attend Tell" is an attention based image captioning model. Some samples are presented in Appendix~\ref{appendix:caption}.

\begin{table}
\small
    \centering
    \caption{Results for image captioning on the MS COCO dataset, the higher the better for all metrics (BLEU 1 to 4, METEOR, and CIDEr). We reimplement both Show\&Tell~\citep{vinyals2015show} and Soft Attention~\citep{xu2015show} in order to add the twin cost. We use two types of images features extracted either with Resnet-101 or Resnet-152.}
    \label{tab:captioning}
    \begin{tabular}{lrrrrrr}
    \toprule
    \textbf{Models} & \textbf{B-1} & \textbf{B-2} & \textbf{B-3} & \textbf{B-4} & \textbf{METEOR} & \textbf{CIDEr} \\
    \midrule
    DeepVS~\citep{karpathy2015deep}  &    62.5 & 45.0 & 32.1 & 23.0 & 19.5 & 66.0 \\
    ATT-FCN~\citep{you2016image} &    70.9 & 53.7 & 40.2 & 30.4 & 24.3 & - \\
    Show \& Tell~\citep{vinyals2015show} & -  & -    & -    & 27.7 & 23.7 & 85.5 \\  
    Soft Attention~\citep{xu2015show}  & 70.7 & 49.2 & 34.4 & 24.3 & 23.9 & - \\
    Hard Attention~\citep{xu2015show}  &  71.8 & 50.4 & 35.7 & 25.0 & 23.0 & - \\
    MSM~\citep{yao2016boosting} & 73.0 & 56.5 & 42.9 & 32.5 & 25.1 & 98.6 \\
    Adaptive Attention~\citep{lu2016knowing}  & {\bf 74.2} & {\bf 58.0} & {\bf 43.9} &  {\bf 33.2} & {\bf 26.6} & {\bf 108.5} \\
    \midrule
    \emph{No attention, Resnet101} \\
    Show\&Tell (Our impl.)           & 69.4 & 51.6 & 36.9 & 26.3 & 23.4 & 84.3\\
     + TwinNet            & {\bf 71.8} & {\bf 54.5} & {\bf 39.4} & {\bf 28.0} & {\bf 24.0} & {\bf 87.7}\\
    \midrule
    \emph{Attention, Resnet101} \\
    Soft Attention (Our impl.) & 71.0 & 53.7 & 39.0 & 28.1 & 24.0 & 89.2\\
    % Soft Attention (Ours) + Sched. Sampl.   & 71.7 & 54.4 & 39.4 & 28.1 & 23.9 & 89.0 \\
    % dima: do we really need scheduled sampling?
    % \; Scheduled Sampling   & 
    % Soft Attention (Ours) + TwinNet (learnt $g$)    & 71.6 & 54.4 & 39.4 & 28.3 & 24.2 &  90.2 \\
    % so far:
    % Soft Attention (Ours, resnet151) & 73.1 & 56.2 & 41.4 & 30.1 & 25.3 & 96.7 \\

     + TwinNet & {\bf 72.8} & {\bf 55.7} & {\bf 41.0} & {\bf 29.7} & {\bf 25.2} & {\bf 96.2} \\
    %Soft Attention + TwinNet^\ref{foot:hyper} & {\bf 71.3} & {\bf 54.1} & {\bf 39.4} & {\bf 28.4} & {\bf 24.3} & {\bf 90.0} \\
    % Bleu_1: 0.713 Bleu_2: 0.542 Bleu_3: 0.394 Bleu_4: 0.284 METEOR: 0.241 ROUGE_L: 0.520  CIDEr: 0.895
    %Epoch 9, stuck Show\&Tell + TwinNet (L2) & 72.9 & 56.0 & 41.3 & 30.2 & 25.3 & 97.1 
    \midrule
    \emph{No attention, Resnet152} \\
    Show\&Tell (Our impl.)           & 71.7 & 54.4 & 39.7 & 28.8 & 24.8 & 93.0\\
     + TwinNet      &  {\bf 72.3} & {\bf 55.2} & {\bf 40.4}  & {\bf 29.3}  & {\bf 25.1} & {\bf 94.7} \\

    \midrule
    \emph{Attention, Resnet152} \\
    Soft Attention (Our impl.) & 73.2 & 56.3 & 41.4 & 30.1 & {\bf 25.3} & 96.6\\
     + TwinNet & {\bf 73.8} & {\bf 56.9} & {\bf 42.0} & {\bf 30.6} & { 25.2} & {\bf 97.3} \\
    \bottomrule
    \end{tabular}
\end{table}

\subsection{Unconditional Generation: Sequential MNIST and Language Modeling}

\sisetup{detect-weight=true,detect-inline-weight=math}
\begin{table}[t]
\centering
\caption{\label{tab:mnist} {\bf (left)} Test set negative log-likelihood for binarized sequential MNIST, where~$^\blacktriangledown$ denotes lower performance of our model with respect to the baselines. {\bf (right)} Perplexity results on WikiText-2 and Penn Treebank~\citep{merity2017regularizing}. AWD-LSTM refers to the model of~\citep{merity2017regularizing} trained with the official implementation at \texttt{http://github.com/salesforce/awd-lstm/}.}
\begin{minipage}[t]{0.49\textwidth}
\small
\begin{tabular}[t]{l
    S[
        table-format=2.2,
        table-space-text-pre={$\approx$}
	]}
\toprule
\textbf{Model} & \textbf{MNIST}\\
\midrule
DBN 2hl{\small ~\citep{germain2015made}} &     {$\approx$}84.55 \\
NADE{\small~\citep{uria2016neural}}  &      88.33 \\
EoNADE-5 2hl{\small~\citep{raiko-nips2014}} &     84.68 \\
DLGM 8 ~\citep{salimans2014markov}       &       {$\approx$}85.51 \\
DARN 1hl~\citep{gregor2015draw} &     {$\approx$}84.13 \\
DRAW~\citep{gregor2015draw}        &      {$\leq$}80.97 \\
P-Forcing$_{(\text{3-layer})}${\small~\citep{lamb2016professor}}     & 79.58 \\
PixelRNN$_{\text{(1-layer)}}${\small~\citep{oord2016pixel}}           & 80.75 \\
PixelRNN$_{\text{(7-layer)}}${\small~\citep{oord2016pixel}}           & 79.20 \\
PixelVAE{\small~\citep{gulrajani2016pixelvae}} & 79.02$^\blacktriangledown$ \\
MatNets{\small~\citep{bachman2016architecture}} & 78.50$^\blacktriangledown$ \\
\midrule
Baseline LSTM$_{\text{(3-layers)}}$ & 79.87 \\
+ TwinNet$_{\text{(3-layers)}}$ & \bfseries 79.35 \\
\midrule
Baseline LSTM$_{\text{(3-layers)}}$ + dropout & 79.59 \\
+ TwinNet$_{\text{(3-layers)}}$ & \bfseries 79.12 \\
\bottomrule
\end{tabular}
\end{minipage}
\begin{minipage}[t]{0.49\textwidth}
\small
\centering
\begin{tabular}[t]{lrr}
\toprule
\textbf{Penn Treebank} & \textbf{Valid} & \textbf{Test} \\
\midrule
%Variational LSTM + Zoneout & 108.7 & 100.9 \\
%Variational LSTM & 101.7 & 96.3 \\
LSTM~\citep{zaremba2014recurrent} & 82.2 & 78.4 \\
4-layer LSTM~\citep{melis2017state} & 67.9 & 65.4 \\
5-layer RHN~\citep{melis2017state} & 64.8 & 62.2 \\
\midrule
AWD-LSTM & 61.2 & 58.8 \\
+ TwinNet & \textbf{61.0} & \textbf{58.3} \\
\bottomrule
%\midrule
%AWD-LSTM & 68.7 & 65.6 \\
%AWD-LSTM + TwinNet & \textbf{68.0} & \textbf{64.9} \\
%\bottomrule
\end{tabular}
\vspace{1mm}\\
\begin{tabular}[b]{lrr}
\toprule
\textbf{WikiText-2} & \textbf{Valid} & \textbf{Test} \\
\midrule
5-layer RHN~\citep{melis2017state} & 78.1 & 75.6 \\
1-layer LSTM~\citep{melis2017state} & 69.3 & 65.9 \\
2-layer LSTM~\citep{melis2017state} & 69.1 & 65.9 \\
\midrule
AWD-LSTM & 68.7 & 65.8 \\
 + TwinNet & \textbf{68.0} & \textbf{64.9} \\
\bottomrule
\end{tabular}
\end{minipage}
\end{table}

% \begin{figure}[t]
% \centering
% \begin{subfigure}[b]{0.45\textwidth}
% \includegraphics[width=\linewidth]{figures/imp4.pdf}
% \caption{\label{fig:mnist75}$75\%$ pixels visible.}
% \end{subfigure}
% \begin{subfigure}[b]{0.45\textwidth}
% \includegraphics[width=\linewidth]{figures/imp3.pdf}
% \caption{\label{fig:mnist25}$25\%$ pixels visible.}
% \end{subfigure}
% \caption{Imputation results for sequential MNIST (occluded pixels are from the bottom of the image). First row is the original image. Second row is the per-pixel $L_t$ loss (brighter indicates higher loss). Third row is the reconstructed image by the forward model. The fourth row is the reconstructed image by the backward model (we plot the per-pixel predictions).}
% \label{fig:mnist}
% \end{figure}

We investigate the performance of our model in pixel-by-pixel generation for sequential MNIST. We follow the setting described by~\citet{lamb2016professor}: we use an LSTM with 3-layers of 512 hidden units for both forward and backward LSTMs, batch size 20, learning rate 0.001 and clip the gradient norms to 5. We use Adam~\citep{kingma2014adam} as our optimization algorithm and we decay the learning rate by half after $5, 10$, and $15$ epochs. Our results are reported at the Table~\ref{tab:mnist} (left). Our baseline LSTM implementation achieves 79.87 nats on the test set. We observe that by adding the TwinNet regularization cost consistently improves performance in this setting by about 0.52 nats. Adding dropout to the baseline LSTM is beneficial. Further gains were observed by adding both dropout and the TwinNet regularization cost. This last model achieves 79.12 nats on test set. Note that this result is competitive with deeper models such as PixelRNN~\citep{oord2016pixel} (7-layers) and PixelVAE~\citep{gulrajani2016pixelvae} which uses an autoregressive decoder coupled with a deep stochastic auto-encoder.

%In order to analyze the lack of improvement, we formulate a task in which we can control the entropy of the output distribution. To this end, we keep $k = 25\%$ (high entropy), $k = 75\%$ (low entropy) of the pixels from the top of the image and let the model to impute the remaining pixels conditioned on the visible pixels (Fig.~\ref{fig:mnist}). Specifically, similarly to the previous conditioned generation experiments, we encode the first $k\%$ of the pixels with the forward RNN and give the last hidden state as a conditioning info to both the backward RNN and the forward RNN. Therefore, the backward RNN is informed about the visible pixels in the digit, which we expect to cause lower entropy in the distribution of backward states. We observe that when $75\%$ of the pixels are visible (Fig.~\ref{fig:mnist75}), TwinNet does not only converges faster but shows better performance on the test set ($11.01$ nats vs $11.15$ nats of the baseline). Results obtained for the $k = 25\%$ case (Fig.~\ref{fig:mnist25}) don't show improvements w.r.t. the baseline. By comparing the second row in Fig.~\ref{fig:mnist}, we note that in the former low-entropy case (Fig.~\ref{fig:mnist75}) regions in which the twin cost is high are regions in which the forward and backward network disagree about the shape of the particular digit. Instead, in the latter high-entropy case (Fig.~\ref{fig:mnist25}), the twin cost appears blurrier,~i.e. less predictive of the possible digits, and is high in regions in which it is unlikely to have ``on'' pixels.

As a last experiment, we report results obtained on a language modelling task using the PennTree Bank and WikiText-2 datasets~\citep{merity2017regularizing}. We augment the state-of-the-art AWD-LSTM model~\citep{merity2017regularizing} with the proposed TwinNet regularization cost. The results are reported in Table~\ref{tab:mnist} (right). 

\section{Discussion}
\label{sec:conclusion}
In this paper, we presented a simple recurrent neural network model that has two separate networks running in opposite directions during training. Our model is motivated by the fact that states of the forward model should be predictive of the entire future sequence. This may be hard to obtain by optimizing one-step ahead predictions. The backward path is discarded during the sampling and evaluation process, which makes the sampling process efficient. Empirical results show that the proposed method performs well on conditional generation for several tasks. The analysis reveals an interpretable behaviour of the proposed loss.

One of the shortcomings of the proposed approach is that the training process doubles the computation needed for the baseline (due to the  backward network training). However, since the backward network is discarded during sampling, the sampling or inference process has the exact same computation steps as the baseline. This makes our approach applicable to models that requires expensive sampling steps, such as PixelRNNs~\citep{oord2016pixel} and WaveNet~\citep{oord2016wavenet}. 
%It would be interesting to show 
One of future work directions is to test whether it could help in conditional speech synthesis using WaveNet.

We observed that the proposed approach yield minor improvements when applied to language modelling with PennTree bank. We hypothesize that this may be linked to the amount of entropy of the target distribution. In these high-entropy cases, at any time-step in the sequence, the distribution of backward states may be highly multi-modal (many possible futures may be equally likely for the same past). One way of overcoming this problem would be to replace the proposed L2 loss (which implicitly assumes a unimodal distribution of the backward states) by a more expressive loss obtained by either employing an inference network~\citep{kingma2013auto} or distribution matching techniques~\citep{goodfellow2014generative}. We leave that for future investigation.

%However, for the conditional case the encoder RNN is much larger than the decoder. Therefore, the additional time for training is negligible. The decoding speed stays the same since the second network is discarded for evaluation.
%The future directions for this work include better theoretical understanding of the proposed method.

\section*{Acknowledgments}
The authors would like to acknowledge the support of the following agencies for research funding and computing support: NSERC, Calcul Qu\'{e}bec, Compute Canada, the Canada Research Chairs, CIFAR, and Samsung. We would also like to thank the developers of Theano~\cite{2016arXiv160502688short}, Blocks and Fuel~\cite{MerrienboerBDSW15}, and Pytorch for developments of great frameworks. We thank Aaron Courville, Sandeep Subramanian, Marc-Alexandre C\^ot\'e, Anirudh Goyal, Alex Lamb, Philemon Brakel, Devon Hjelm,   Kyle Kastner, Olivier Breuleux, Phil Bachman, and Ga\'etan Marceau Caron  for useful feedback and discussions.

\vfill \pagebreak
\bibliography{iclr2018_conference}
\bibliographystyle{iclr2018_conference}

%\appendix\section{Appendix: Image captioning samples}
%\label{appendix:caption}
% TODO: add images and samples for captioning

\end{document}